\documentclass[runningheads]{llncs}
\usepackage{amsmath,amssymb,amsfonts,dsfont,bm,mathrsfs}
\usepackage{graphicx}

\newcommand{\argmax}{\mathop{\arg\max}}  

\newcommand{\argmin}{\mathop{\arg\min}}

\begin{document}

\title{Interpreting Generative Adversarial Networks for Interactive Image Generation}

\titlerunning{Interpreting Generative Adversarial Networks}



%

\author{Bolei Zhou\orcidID{0000-0003-4030-0684}}

\authorrunning{Bolei Zhou}



%

\institute{Department of Computer Science, University of California, Los Angeles}
\maketitle              
\begin{abstract}

Significant progress has been made by the advances in Generative Adversarial Networks (GANs) for image generation. However, there lacks enough understanding of how a realistic image is generated by the deep representations of GANs from a random vector. This chapter gives a summary of recent works on interpreting deep generative models. The methods are categorized into the supervised, the unsupervised, and the embedding-guided approaches. We will see how the human-understandable concepts that emerge in the learned representation can be identified and used for interactive image generation and editing. 


\end{abstract}
\section{Introduction}
\begin{figure}
  \centering
  \includegraphics[width=1.0\linewidth]{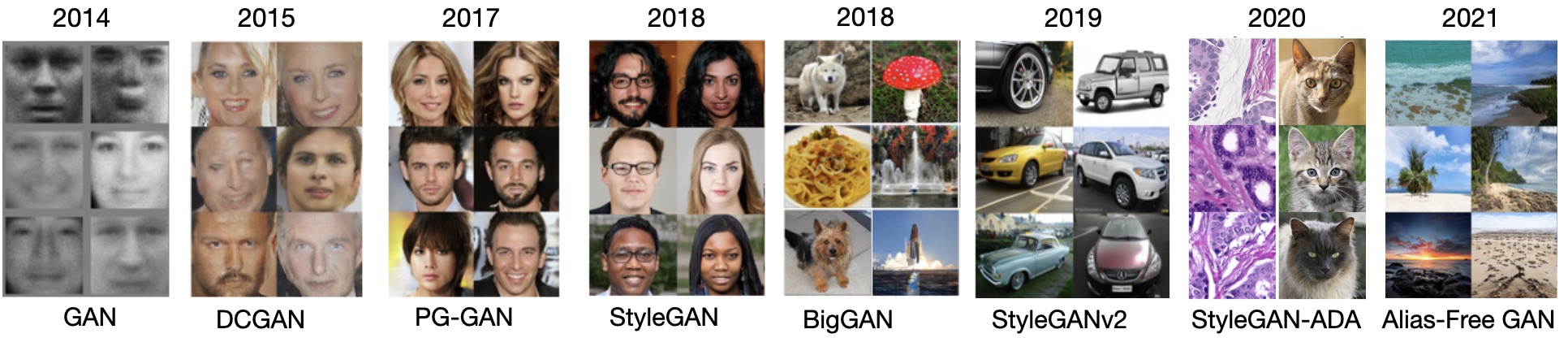}
  \caption{Progress of image generation made by different GAN models over the years.}\label{fig:progress}
\end{figure}

Over the years, great progress has been made in image generation by the advances in Generative Adversarial Networks (GANs)~\cite{karras2019style,goodfellow2014generative}. As shown in Fig.\ref{fig:progress} the generation quality and diversity have been improved substantially from the early DCGAN~\cite{radford2015unsupervised} to the very recent Alias-free GAN~\cite{aliasfreegan}. After the adversarial training of the generator and the discriminator, we can have the generator as a pretrained feedforward network for image generation. After feeding a vector sampled from some random distribution, this generator can synthesize a realistic image as the output. However, such an image generation pipeline doesn't allow users to customize the output image, such as changing the lighting condition of the output bedroom image or adding a smile to the output face image. Moreover, it is less understood how a realistic image can be generated from the layer-wise representations of the generator. Therefore, we need to interpret the learned representation of deep generative models for understanding and the practical application of interactive image editing. 

This chapter will introduce the recent progress of the interpretable machine learning for deep generative models. I will show how we can identify the human-understandable concepts in the generative representation and use them to steer the generator for interactive image generation. Readers might also be interested in watching a relevant tutorial talk I gave at CVPR'21 Tutorial on Interpretable Machine Learning for Computer Vision\footnote{https://youtu.be/PtRU2B6Iml4}. A more detailed survey paper on GAN interpretation and inversion can be found in \cite{xia2021gan}. 

This chapter focuses on interpreting the pretrained GAN models, but a similar methodology can be extended to other generative models such as VAE. Recent interpretation methods can be summarized into the following three approaches: the supervised approach, the unsupervised approach, and the embedding-guided approach. The supervised approach uses labels or classifiers to align the meaningful visual concept with the deep generative representation; the unsupervised approach aims to identify the steerable latent factors in the deep generative representation through solving an optimization problem; the embedding-guided approach uses the recent pretrained language-image embedding CLIP~\cite{radford2021learning} to allow a text description to guide the image generation process.

In the following sections, I will select representative methods from each approach and briefly introduce them as primers for this rapidly growing direction. 

\section{Supervised Approach}
\begin{figure}
  \centering
  \includegraphics[width=1.0\linewidth]{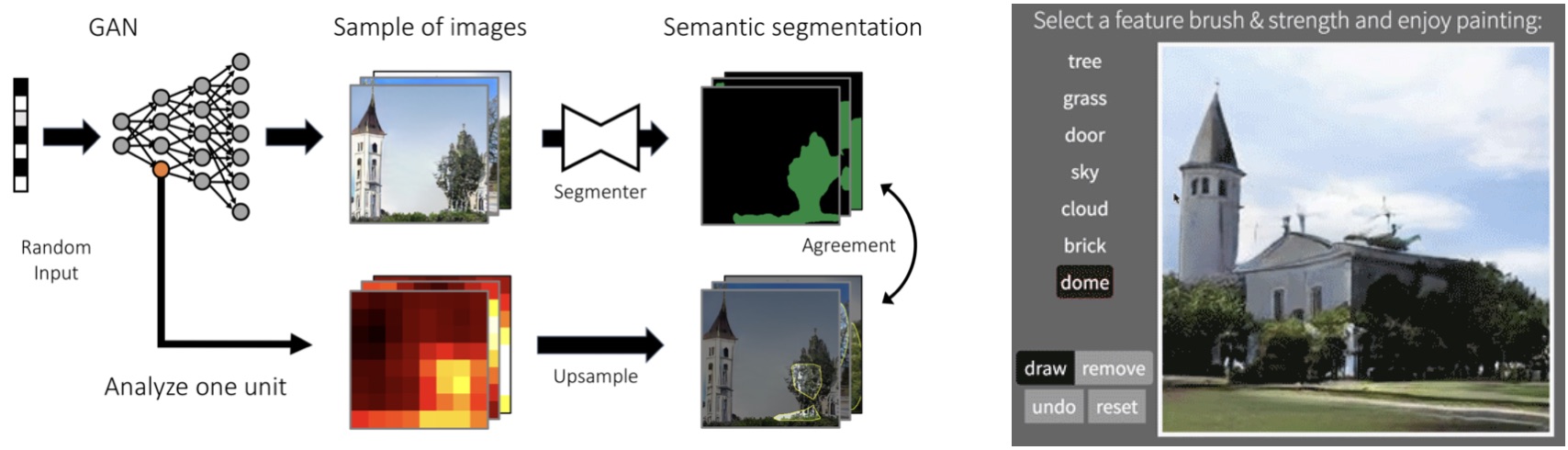}
  \caption{GAN dissection framework and interactive image editing interface. Images are extracted from \cite{bau2017network}. The method aligns the unit activation with the semantic mask of the output image, thus by turning up or down the unit activation we can include or remove the corresponding visual concept in the output image. }\label{fig:gandissect}
\end{figure}

The supervised approach uses labels or trained classifiers to probe the representation of the generator. One of the earliest interpretation methods is the GAN Dissection~\cite{bau2018gan}. Derived from the previous work Network Dissection~\cite{bau2017network}, GAN Dissection aims to visualize and understand the individual convolutional filters (we term them as units) in the pretrained generator. It uses semantic segmentation networks~\cite{zhou2017scene} to segment the output images. It then calculates the agreement between the spatial location of the unit activation map and the semantic mask of the output image. This method can identify a group of interpretable units closely related to object concepts, such as sofa, table, grass, buildings. Those units are then used as switches where we can add or remove some objects such as a tree or lamp by turning up or down the activation of the corresponding units. The framework of GAN Dissection and the image editing interface are shown in Fig.\ref{fig:gandissect}. In the interface of GAN Dissection, the user can select the object to be manipulated and brush the output image where it should be removed or added. 

Besides steering the filters at the intermediate convolutional layer of the generator as the GAN Dissection does, the latent space where we sample the latent vector as input to the generator is also being explored. The underlying interpretable subspaces aligning with certain attributes of the output image can be identified. Here we denote the pretrained generator as $G(.)$ and the random vector sampled from the latent space as $\textbf{z}$, and then the output image becomes $I = G(\textbf{z})$. Under different vectors, the output images become different. Thus the latent space encodes various attributes of images. If we can steer the vector $\textbf{z}$ through one relevant subspace and preserve its projection to the other subspaces, we can edit one attribute of the output image in a disentangled way. 

\begin{figure}
  \centering
  \includegraphics[width=1.0\linewidth]{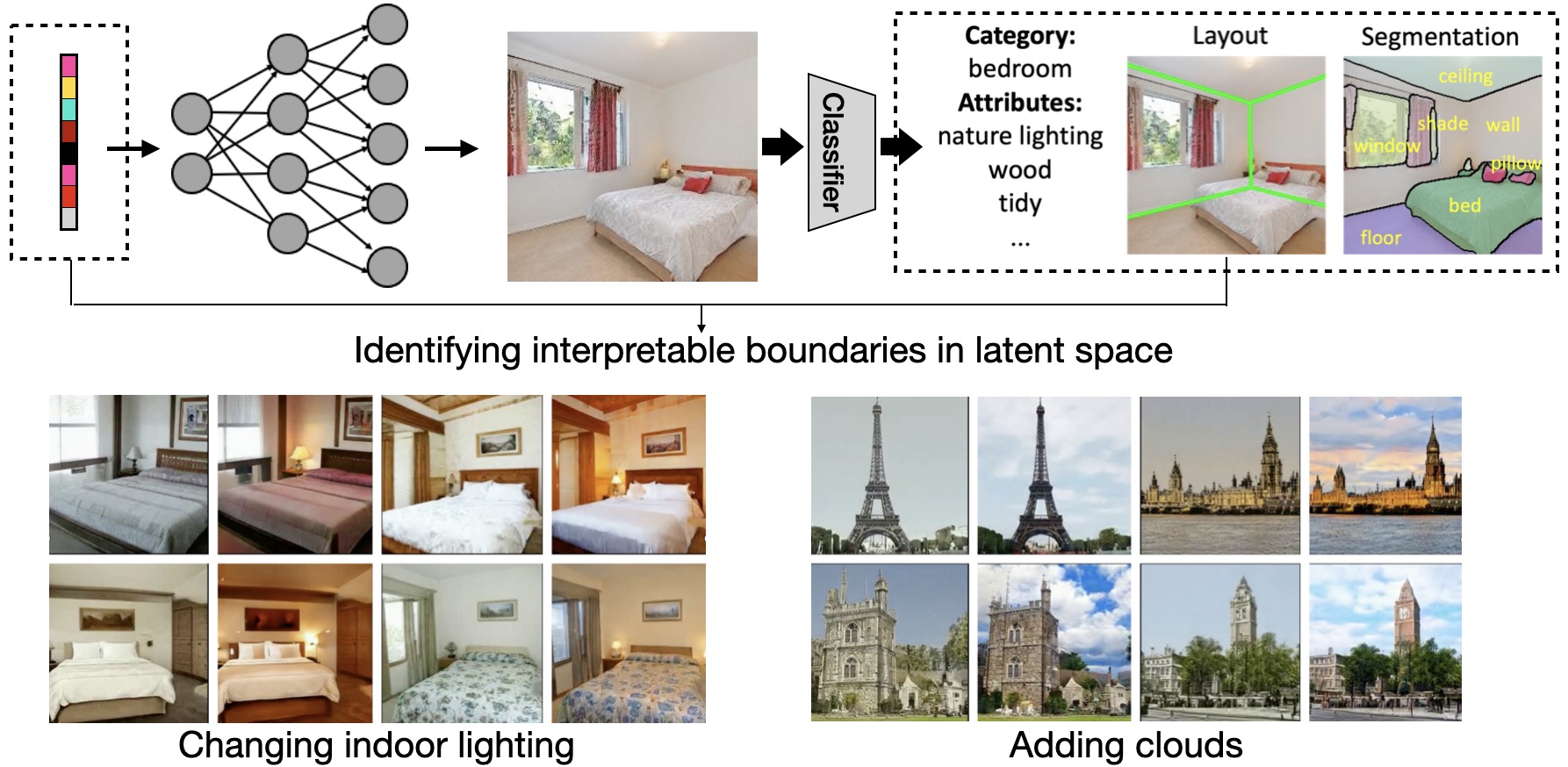}
  \caption{We can use classifier to predict various attributes from the output image then go back to the latent space to identify the attribute boundaries. Images below show the image editing results achieved by \cite{yang2021semantic}.}\label{fig:higan}
\end{figure}

To align the latent space with the semantic space, we can first apply off-the-shelf classifiers to extract the attributes of the synthesized images and then compute the causality between the occurring attributes in the generated images and the corresponding vectors in the latent space. The HiGAN method proposed in \cite{yang2021semantic} follows such a supervised approach as illustrated in Fig.\ref{fig:higan}: (1) Thousands of latent vectors are sampled, and the images are generated. (2) Various levels of attributes are predicted from the generated images by applying the off-the-shelf classifiers. (3) For each attribute $a$, a linear boundary $\textbf{n}_a$ is trained in the latent space using the predicted labels and the latent vectors. We consider it a binary classification and train a linear SVM to recognize each attribute. The weight of the trained SVM is $\textbf{n}_a$. (4) a counterfactual verification step is taken to pick up the reliable boundary. Here we follow a linear model to shift the latent code as

\begin{equation}
    I' = G(\textbf{z} + \lambda\textbf{n}_a),
\end{equation}

where the normal vector of the trained attribute boundary is denoted as $\textbf{n}_a$ and $I'$ is the edited image compared to the original image $I$. Then the difference between predicted attribute scores before and after manipulation becomes, 

\begin{equation}
    \Delta a = \frac{1}{K}\sum_{k=1}^K \max(F(G(\textbf{z}_k + \textbf{n}_a)) - F(G(\textbf{z}_k)), 0),
\end{equation}

here $F(.)$ is the attribute predictor with the input image, and $K$ is the number of synthesized images. Ranking $\Delta a$ allows us to identify the reliable attribute boundaries out of the candidate set $\{ \textbf{n}_a\}$, where there are about one hundred attribute boundaries trained from step 3 of the HiGAN method. After that, we can then edit the output image from the generator by adding or removing the normal vector of the target attribute on the original latent code. Some image manipulation results are shown in Fig.\ref{fig:higan}. 

Similar supervised methods have been developed to edit the facial attributes~\cite{shen2020interpreting,shen2020interfacegan} and improve the image memorability~\cite{goetschalckx2019ganalyze}. Steerability of various attributes in GANs has also been analyzed~\cite{jahanian2019steerability}. Besides, the work of StyleFlow~\cite{abdal2021styleflow} replaces the linear model with a nonlinear invertible flow-based model in the latent space with more precise facial editing. Some recent work uses a differentiable renderer to extract 3D information from the image GANs for more controllable view synthesis~\cite{zhang2020image}. For the supervised approach, many challenges remain for future work, such as expanding the annotation dictionary, achieving more disentangled manipulation, and aligning latent space with image region. 

\section{Unsupervised Approach}
As generative models become more and more popular, people start training them on a wide range of images, such as cats and anime. To steer the generative models trained for cat or anime generation, following the previous supervised approach, we have to define the attributes of the images and annotate many images to train the classifiers. It is a very time-consuming process. 

Alternatively, the unsupervised approach aims to identify the controllable dimensions of the generator without using labels/classifiers. 

\begin{figure}
  \centering
  \includegraphics[width=1.0\linewidth]{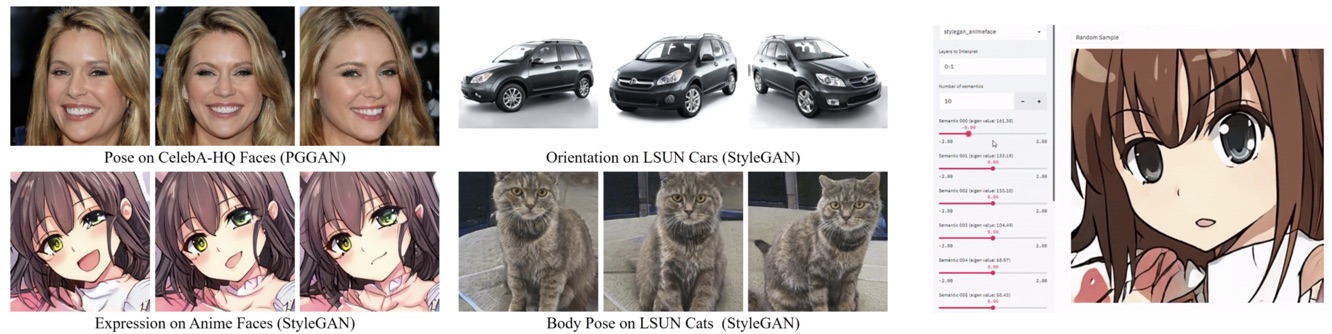}
  \caption{Manipulation results from SeFa~\cite{shen2021closed} on the left and the interface for interactive image editing on the right. On the left, each attribute corresponds to some $\textbf{n}_i$ in the latent space of the generator. In the interface, user can simply drag each slider bar associating with certain attribute to edit the output image}\label{fig:sefa}
\end{figure}

SeFa~\cite{shen2021closed} is an unsupervised approach for discovering the interpretable representation of a generator. It directly decomposes the pre-trained weights. More specifically, in the pre-trained generator of the popular StyleGAN~\cite{karras2019style} or PGGAN~\cite{karras2017progressive} model, there is an affine transformation between the latent code and the internal activation. Thus the manipulation model can be simplified as
\begin{align}
  \textbf{y}' \triangleq G_1(\textbf{z}') &= G_1(\textbf{z} + \alpha\textbf{n}) \nonumber \\
    &= \textbf{A}\textbf{z} + \textbf{b} + \alpha\textbf{A}\textbf{n} = \textbf{y} + \alpha\textbf{A}\textbf{n},
             \label{eq:new-manipulation}
\end{align}

where $\textbf{y}$ is the original projected code and $\textbf{y}'$ is the projected code after manipulation by \textbf{n}. From Eq.~\eqref{eq:new-manipulation} we can see that the manipulation process is instance independent.
In other words, given any latent code \textbf{z} together with a particular latent direction \textbf{n}, the editing can always be achieved by adding the term $\alpha\textbf{A}\textbf{n}$ onto the projected code after the first step.
From this perspective, the weight parameter $\textbf{A}$ should contain the essential knowledge of the image variation. Thus we aim to discover important latent directions by decomposing $\textbf{A}$ in an unsupervised manner. We propose to solve the following optimization problem:
\begin{align}
  \textbf{N}^* = \argmax_{\{\textbf{N}\in R^{d\times k}: \textbf{n}_i^T\textbf{n}_i = 1\ \forall i=1,\cdots,k\}} \sum_{i=1}^k ||\textbf{A}\textbf{n}_i||_2^2,  
\end{align}
where $\textbf{N} = [\textbf{n}_1, \textbf{n}_2, \cdots, \textbf{n}_k]$ correspond to the top $k$ semantics sorted by their eigenvalues, and $\textbf{A}$ is the learned weight in the affine transform between the latent code and the internal activation. This objective aims at finding the directions that can cause large variations after the projection of $\textbf{A}$. The resulting solution becomes the eigenvectors of the matrix $\textbf{A}^T\textbf{A}$. Those resulting directions at different layers control different attributes of the output image, thus pushing the latent code \textbf{z} on the  important directions $\{\textbf{n}_1, \textbf{n}_2, \cdots, \textbf{n}_k\}$ facilitates the interactive image editing.  Fig.\ref{fig:sefa} shows some editing result.

Many other methods have been developed for the unsupervised discovery of interpretable latent representation. H\"{a}rk\"{o}nen \textit{et al.}~\cite{harkonen2020ganspace} perform PCA on the sampled data to find primary directions in the latent space. Voynov and Babenko~\cite{voynov2020unsupervised} jointly learn a candidate matrix and a classifier such that the classifier can properly recognize the semantic directions in the matrix. Peebles \textit{et al.}~\cite{peebles2020hessian} develops a Hessian penalty as a regularizer for improving disentanglement in training. He \textit{et al.}~\cite{he2021eigengan} designs a linear subspace with an orthogonal basis in each layer of the generator to encourage the decomposition of attributes. Many challenges remain for the unsupervised approach, such as how to evaluate the result from unsupervised learning, annotate each discovered dimension, and improve the disentanglement in the GAN training process. 
\section{Embedding-guided Approach}

The embedding-guided approach aligns language embedding with generative representations. It allows users to use any free-form text to guide the image generation. The difference between the embedding-guided approach and the previous unsupervised approach is that the embedding-guided approach is conditioned on the given text to manipulate the image to be more flexible, while the unsupervised approach discovers the steerable dimensions in a bottom-up way thus it lacks fine-grained control. 

\begin{figure}
  \centering
  \includegraphics[width=1.0\linewidth]{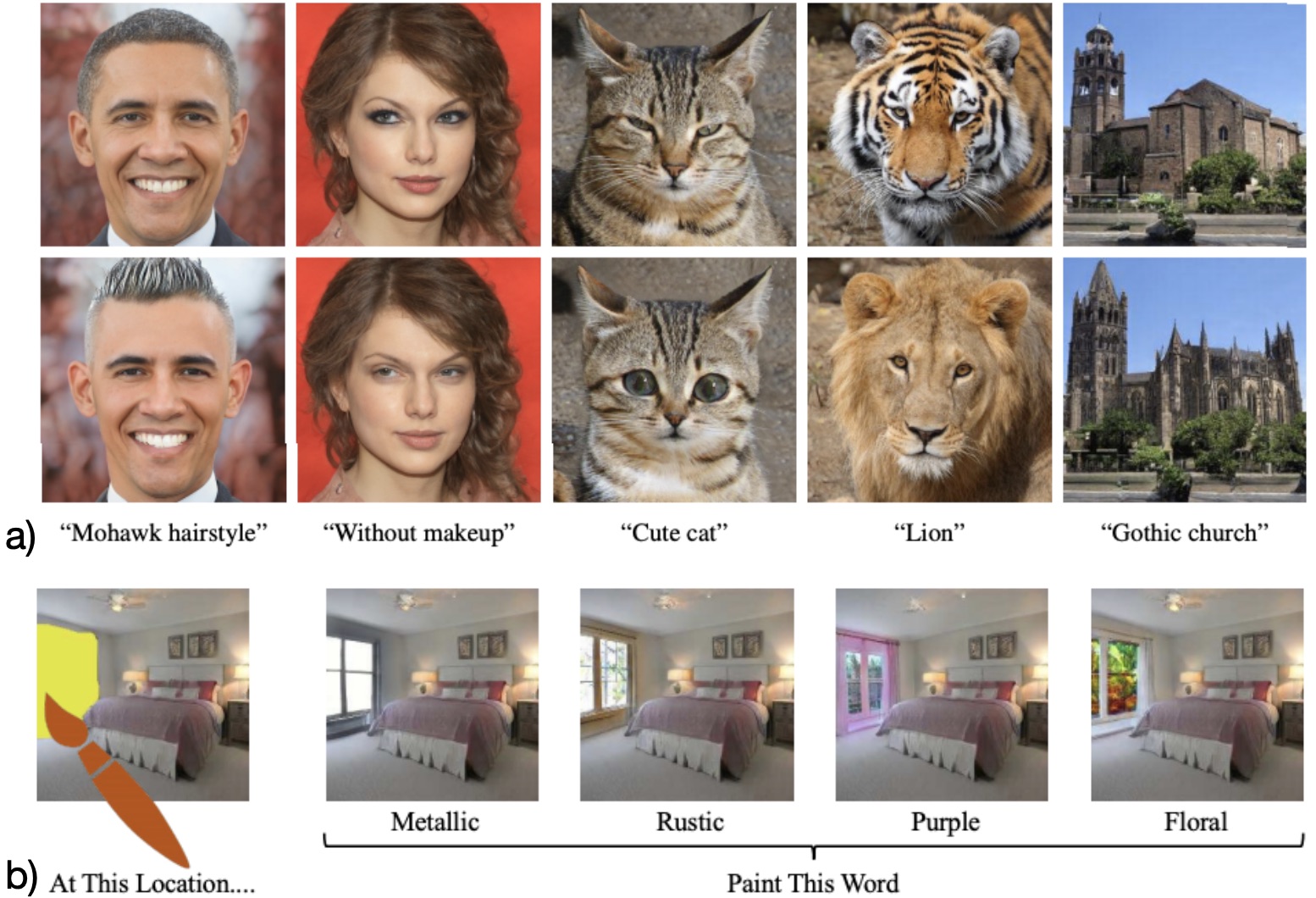}
  \caption{Text driven image editing results from a) StyleCLIP~\cite{patashnik2021styleclip} and b) Paint by Word~\cite{bau2021paint}. }\label{fig:zeroshot}
\end{figure}

Recent work on StyleCLIP~\cite{patashnik2021styleclip} combines the pretrained language-image embedding CLIP~\cite{radford2021learning} and StyleGAN generator~\cite{karras2019style} for free-form text-driven image editing. CLIP is a pretrained embedding model from 400 million image-text pairs. Given an image $I_s$, it first projects it back into the latent space as $\textbf{w}_s$ using existing GAN inversion method. Then StyleCLIP designs the following optimization objective 

\begin{equation}
    \textbf{w}^* = \argmin D_{CLIP}(G(\textbf{w}), t) + \lambda_{L2}||\textbf{w} - \textbf{w}_s||_2 + \lambda_{ID}L_{ID}(\textbf{w}, \textbf{w}_s),
\end{equation}

where $D_{CLIP}(.,.)$ measure the distance between an image and a text using the pre-trained CLIP model, the second and the third terms are some regularizers to keep the similarity and identity with the original input image. Thus this optimization objective results in a latent code $\textbf{w}^*$ that generates an image close to the given text in the CLIP embedding space as well as similar to the original input image. StyleCLIP further develops some architecture design to speed up the iterative optimization. Fig.\ref{fig:zeroshot} shows the text driven image editing results. 

Some concurrent work called Paint by Word from Bau \textit{et al.}~\cite{bau2021paint} combines CLIP embedding with region-based image editing. It has a masked optimization objective that allows the user to brush the image to provide the input mask. 

\section{Concluding Remarks}

Interpreting deep generative models leads to a deeper understanding of how the learned representations decompose images to generate them. Discovering the human-understandable concepts and steerable dimensions in the deep generative representations also facilitates the promising applications of interactive image generation and editing. We have introduced representative methods from three approaches: the supervised approach, the unsupervised approach, and the embedding-guided approach. The supervised approach can achieve the best image editing quality when the labels or classifiers are available. It remains challenging for the unsupervised and embedding-guided approaches to achieve disentangled manipulation. More future works are expected on the accurate inversion of the real images and the precise local and global image editing. 

\bibliographystyle{splncs04}

\bibliography{sample}

\end{document}